# Exploring Lexical, Syntactic, and Semantic Features for Chinese Textual Entailment in NTCIR RITE Evaluation Tasks

Wei-Jie Huang[†] and Chao-Lin Liu[‡]

[†‡] Department of Computer Science, National Chengchi University, Taiwan

[‡] Graduate Institute of Linguistics, National Chengchi University, Taiwan

[†]s951553@gmail.com, [‡]chaolin@nccu.edu.tw

**Abstract**

We computed linguistic information at the lexical, syntactic, and semantic levels for Recognizing Inference in Text (RITE) tasks for both traditional and simplified Chinese in NTCIR-9 and NTCIR-10. Techniques for syntactic parsing, named-entity recognition, and near synonym recognition were employed, and features like counts of common words, statement lengths, negation words, and antonyms were considered to judge the entailment relationships of two statements, while we explored both heuristics-based functions and machine-learning approaches. The reported systems showed robustness by simultaneously achieving second positions in the binary-classification subtasks for both simplified and traditional Chinese in NTCIR-10 RITE-2. We conducted more experiments with the test data of NTCIR-9 RITE, with good results. We also extended our work to search for better configurations of our classifiers and investigated contributions of individual features. This extended work showed interesting results and should encourage further discussion.

**Keywords**: Textual entailment recognition, Negation and antonyms, Near synonym recognition, Named-entity recognition, Dependency parsing, Trained heuristic functions, Support-vector machines, Linearly weighted models, Decision trees

## 1 Introduction[1]

Recognizing textual entailment (RTE) (Dagan, Glickman, & Magnini 2006) has become a major research topic in natural language processing (NLP) in the past decade (Watanabe et al. 2013). Given a pair of statements, *text* (*T*) and *hypothesis* (*H*), the most basic format of an RTE task is to determine whether *H* is true when *T* is true; namely, whether or not *T* entails *H*. A more challenging format is to determine whether *T* and *H* are contradictory statements (Dagan, Dolan, Magnini, & Roth 2009). More recently in PASCAL RTE-6[2], NTCIR-10 RITE-2[3], and NTCIR-11 RITE-VAL[4], researchers investigated and evaluated methods for identifying statements in a collection, e.g., a corpus like Wikipedia, which are relevant to a given statement *T*, where relevancy includes both entailment, paraphrase, and contradiction.

---

[1] The pronunciations and translations of all Chinese strings mentioned in this paper are provided in the Appendix.
[2] http://www.nist.gov/tac/2010/RTE/
[3] http://www.cl.ecei.tohoku.ac.jp/rite2
[4] https://sites.google.com/site/ntcir11riteval/

The RTE tasks are relevant and applicable to many NLP applications, including knowledge management (Tsujii 2012). If a statement entails another in a collection of statements, then one may not need to consider both statements to produce a concise summary of the collection, so recognizing entailments is useful for automatic text summarization (Lloret, Ferrández, Muñoz, & Palomar 2008; Tatar, Mihis, Lupsa, & Tamaianu-Morita 2009). Similar reasons apply to how recognizing entailment can be applied to question answering systems (de Salvo Braz et al. 2005). When a question entails another, the recorded answer to the previous question may be useful for answering the new question. RTE can also be useful for judging the correctness of students' descriptive answers in assessment tasks. It is rare for students to respond to questions with statements that are exactly the same as the instructors' standard answers. It is also not practical to expect instructors to list all possible ways which students may answer a question. In such cases, recognizing paraphrase relationships between students' and instructors' answers becomes instrumental (Nielsen, Ward, & Martin 2009). We have also applied RTE techniques to enable computers to take reading comprehension tests that are designed for middle school students (Huang, Lin, & Liu 2013).

Dagan et al. (2009) provided an overview of the approaches for RTE. Treating RTE as a classification task is an obvious option, where different systems consider various factors to make the final decisions. Due to the availability of the training data in RTE activities, machine learning-based approaches are common. Researchers design methods to utilize different levels of linguistic, including syntactic and semantic, information provided in the given statement pairs to judge their relationships. Transformation-based methods offer interesting alternatives for the RTE tasks. If a statement can be transformed into another via either syntactic rewriting (Bar-Haim et al. 2008; Stern at al. 2011; Shibata, Kurohashi, Kohama, & Yamamoto 2013) or logical inference procedures (Chambers et al. 2007; Takesue & Ninomiya 2013; Wang, Zhao, & Lu 2013; Watanabe, Mizuno, & Inui 2013), then the statements may be highly related. In addition to using the information conveyed by the given statements, external information like common sense knowledge and ontology about problem domains can strengthen the basis on which entailment decisions are made (de Salvo Braz et al. 2005; Stern et al. 2010).

The first corresponding event of PASCAL RTE for Japanese and Chinese took place in NTCIR-9, and was named RITE as the acronym for "**R**ecognizing **I**nference in **TE**xt" (Shima et al. 2012). NTCIR-10 continued to host RITE-2 for Japanese and Chinese, and had, respectively, ten and nine teams participating in the traditional and simplified Chinese subtasks (Watamabe et al. 2013). All of these participants considered different combinations of linguistic information as features to determine the entailment relationships of statement pairs. Most of them employed support vector machines as the classifiers.

There were different subtasks in NTCIR-9 RITE and NTCIR-10 RITE-2. The **binary classification** (**BC**) subtask required participants to judge whether or not *T* entails *H*. In this paper, we will focus only on the BC subtasks in the NTCIR RITE tasks, as we believe that the BC subtask is the most fundamental subtask of them all.

In NTCIR-10 RITE-2, the best performing team in the BC subtask for traditional Chinese (**CT**) adopted a voting mechanism (Shih et al. 2013). The best performing team in the BC subtask for simplified Chinese (**CS**) employed an alignment-based strategy (Wang, Zhao, & Lu 2013). We (Huang & Liu 2013) trained heuristic functions to achieve second best performance in the BC subtasks for both CT and CS. The best team outperformed us in the BC subtask for CT by only 0.7% in the F1 measure. Chang et al. (2013) embraced decision trees as the classifier but did not achieve an impressive performance.

For obvious reasons, all participating systems in NTCIR-10 RITE-2 used some forms of linguistic features to make decisions. As may be expected, different systems considered different sets of features and applied them in different ways. We computed lexical, syntactic, and semantics information about the statement pairs to judge their entailment relationships. The linguistic features were computed with public tools and machine-readable dictionaries, including the Extended HowNet[5] (Chen et al. 2010). Preprocessing steps for the statements included conversion between simplified and traditional Chinese, Chinese segmentation, and converting formats of Chinese numbers. We employed such linguistic information as (1) words that were shared by both statements, (2) synonyms, antonyms, and negation words, (3) information about the named entities of the statement pairs, and (4) similarity between parse trees and dependency structures, etc.

The performance of our approaches was sufficiently robust that we achieved the second best scores in both CT and CS subtasks. Since each participating team could submit running results of three different configurations, we actually experimented with our models that we built by training heuristic functions and support vector machines (SVMs). Our best results were achieved by the trained heuristic functions, achieving second position in the BC subtasks for both CT and CS. Our SVM-based models achieved the third best score in the BC subtask for CT, but dropped to 12[th] position in BC subtask for CS.

We have extended our work after participation in NTCIR-10 RITE-2. We ran grid searches of larger scales to find the best combinations of parameters and features for the classification models. In general, conducting the grid search helped us build better models. However, the experimental results also provide interesting and seemingly perplexing material for further discussion in the paper. We also tested our systems with the test data for the BC subtasks of NTCIR-9 RITE, and found that we were able to achieve better performance than the best performer in NTCIR-9 RITE tasks.

---

[5] http://ehownet.iis.sinica.edu.tw/

We explain the preprocessing of the text material and extraction of their linguistic features in Sect. 2, examine the constructions of the heuristics-based and machine learning-based classifiers in Sect. 3, present and discuss the experimental results in Sect. 4, review and deliberate on some additional observations in Sect. 5, and wrap up this paper in Sect. 6.

## 2  Major System Components

In this section, we describe components of our running systems, including the preprocessing steps and the extraction of fundamental linguistic features.

### 2.1  Preprocessing

In this subsection, we explain the preprocessing steps: traditional-to-simplified Chinese conversion, numeric format conversion, and Chinese segmentation.

#### 2.1.1  Traditional-to-simplified Chinese conversion

We relied on Stanford NLP tools[6] to do Chinese segmentation and named-entity recognition. As those tools were designed to perform better for simplified Chinese we had to convert traditional Chinese into simplified Chinese. We converted words between their traditional and simplified forms of Chinese with an automatic procedure which relied on a tool in Microsoft Word. We did not design or invent a conversion dictionary of our own, and the quality of conversion depended solely on Microsoft Word.

There are two major methods for converting between traditional and simplified Chinese text. The simpler option is just to do character-to-character conversion, e.g., changing "電腦軟體的品質很重要"[7] to "电脑软体的品质很重要". A more sophisticated and better conversion is to do word-to-word conversion, changing this sample statement to "计算机软件的质量很重要". This latter conversion includes the simplified Chinese words, i.e., "计算机", "软件", and "质量" that are used in the training of the Stanford tools, so is more likely to lead to better system performance. Microsoft Word offers the second type of conversion as much as it can, and we understand that Microsoft Word might not convert all traditional Chinese words perfectly to their simplified counterparts, e.g., the result of converting "工業技術水準" is "工业技术水准". "工业技术水平" is a preferred conversion. However, Microsoft Word is a good and accessible current choice.

#### 2.1.2  Numeric format conversion

There are multiple ways for people to write numbers in English text, e.g., sixteen vs. 16. In Chinese, there are at least three ways to write numbers in text, e.g., "3", "三", and "参" for the number "3". There are also specific characters to express specific numbers, e.g., "廿" and "卅" for 20 and 30, respectively. In addition, there are simplified ways to express relatively small numbers, e.g., "三十二" for 32 but "十二" for 12. In the latter case, "一十二" is more formal but is rarely used.

---

[6] http://nlp.stanford.edu/software/index.shtml
[7] The pronunciations and translations of all Chinese strings mentioned in this paper are provided in the Appendix.

To streamline our handling of numbers in Chinese statements, we employed regular expressions to capture specific strings and convert them to Arabic numerals. The conversions need special care for some extraordinary instances. For instance, one may not want to convert "朝九晚五" to "朝 9 晚 5" or convert "舉一反三" to "舉 1 反 3".

### 2.1.3 Chinese string segmentation

We employed the Stanford Word Segmenter[8] (Chang, Galley, & Manning 2008) to segment Chinese character strings into word tokens. Unlike most alphabetical languages in which words are separated by spaces, Chinese text strings do not have delimiters between words. In fact, Chinese text did not use punctuation marks until modern times. In the field of natural language processing, converting a Chinese string into a sequence of Chinese words is called segmentation (or tokenization) of Chinese.

A major challenge of Chinese segmentation is that different segmentations of a given Chinese string can represent very different meanings of the original string. We can segment the string "研究生命還有多少年" in two different ways: {"研究生命", "還有", "多少年"} or {"研究生", "命", "還有", "多少年" }. Adopting the former segmentation, the translation of the original Chinese string is "how many more years can one do research". Adopting the latter will lead to "how many more years can the graduate student survive". To most native speakers of Chinese, the former segmentation is much more natural, but the latter is not unacceptable. In the 2012 Bakeoff for Chinese segmentation, the best performing system reached an F1 measure slightly shy of 95% (Duan, Sui, Tian, & Li 2012)

## 2.2 Lexical semantics

### 2.2.1 Lexical resources and computation for Chinese synonyms

The number of words shared by statement pairs is the most commonly used feature to judge entailment. Identifying words that are shared literally is a direct way to compute word overlaps. Indeed, in previous RTE and RITE events, organizers provided baseline systems which calculated character overlaps to determine entailment (Bar-Haim et al. 2006; Shima et al. 2011).

In practice, people may express the same or very similar ideas with synonyms and near synonyms, so their identification is also very important. The following statements are very close in meaning though they do not use exactly the same words.

---

[8] http://nlp.stanford.edu/software/segmenter.shtml

(1) Tamara is **reluctant** to **raise** this question.

(2) Tamara **hesitates** to **ask** this question.

Translating this pair into Chinese will also show the importance of identifying synonyms.

(3) Tamara 對於**提出**這一問題感到**猶豫**

(4) Tamara 對於**詢問**這一問題顯得**遲疑**

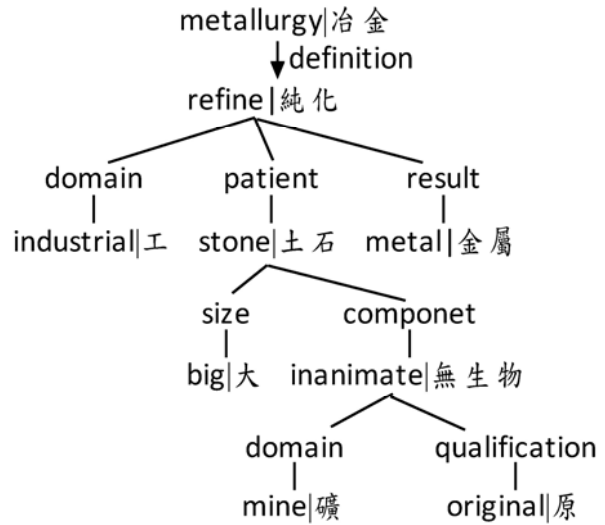

**Fig. 1** A definition tree for "冶金" (metallurgy)

The literature has seen abundant ways to compute synonyms for English, particularly those that computed the similarity between words based on WordNet[9] (Budanitsky & Hirst 2006). In contrast, we have yet to find a good way to compute synonyms for Chinese.

To compute synonyms for a given word, we rely on both existing lexicons and computing methods. We acquired a dictionary for synonyms and antonyms[10] from the Ministry of Education (MOE) of Taiwan. This MOE dictionary lists 16,005 synonyms and 8625 antonyms.

We could employ the extended HowNet[11] (E-HowNet), which can be considered as an extended WordNet for Mandarin Chinese, to look up synonyms of Chinese words. E-HowNet contains 88,079 traditional Chinese words in the 2012 version, and can provide synonyms of Chinese words, so we could use the list of synonyms directly. We will find 38 synonymous words[12] which carry the concept of "hesitate" in E-HowNet. In this particular case, we would be able to tell that "猶豫" in statement (3) and "遲疑" in statement (4) are synonymous with the list in E-HowNet. However, "提出" in statement (3) does not belong to the synonym list[13] of "詢問" in statement (4). "提出" is similar to "raise" in English. One can raise a question or a concern, so "raise" alone does not necessarily relate to asking questions.

We could also use the definitions for words in E-HowNet to estimate the relatedness between two Chinese words by their taxonomical relations and semantic relations (Chuang,

---

[9] http://wordnet.princeton.edu/

[10] http://dict.revised.moe.edu.tw/

[11] http://ehownet.iis.sinica.edu.tw/

[12] 三心二意, 心猿意馬, 彷徨, 投鼠忌器, 沈吟, 沉吟, 委決不下, 狐疑不決, 首鼠兩端, 動搖, 徘徊不前, 逡巡, 游移, 猶猶豫豫, 猶疑, 猶疑不決, **猶豫**, 猶豫不決, 搖擺不定, 當斷不斷, 滯足不前, 裹足, 裹足不前, 踟躕, 踟, 踟躕, 踟躕不前, 踟躕不進, **遲疑**, 遲疑不決, 舉棋不定, 舉棋未定, 瞻前顧後, 躊躇, 顧忌, 躕, 觀望不前, 搖擺

[13] 叩問, 打探, 打聽, 扣問, 咨, 咨詢, 查詢, 討教, 追問, 問, 問到, 問津, 問訊, 問起, 問話, 問道, 探問, 探詢, 尋問, 提問, 敢問, 發問, 詰問, 詢, **詢問**, 詢答, 徵詢, 請旨, 請教, 質詢, 諮諮, 詢諮, 諏訊

Liu, & Chang 2012; Chen 2013; Huang & Liu 2013). In this work, we converted the definition of a word into a "definition tree", e.g., Fig. 1, according to the taxonomy in E-HowNet. Each node represents a primitive unit, a function word, or a semantic role. Considering each internal node in a definition tree as a root, we built a collection of subtrees of the definition tree. In Fig. 1, there are 15 nodes.

The DICE coefficient[14] between the collections of subtrees of two definition trees is used to measure the degree of relatedness of two definitions. Given two collections, e.g., *X* and *Y*, the DICE coefficient is defined in Eq. (1), where |*X*| is the number of elements in *X*.

$$DICE(X,Y) = \frac{2|X \cap Y|}{|X|+|Y|} \qquad (1)$$

Due to the definition, a DICE coefficient must fall in the range of [0, 1]. Two definitions will be considered anonymous if their DICE coefficient is larger than a threshold, for which we chose to use 0.88 based on a small-scale experiment.

Computing Chinese synonyms only with information in dictionaries is an imperfect method. Chinese text contains out-of-vocabulary (OOV) words a lot more frequently than English text. For these OOV words, dictionary-based methods cannot always help.

### 2.2.2 Chinese antonyms and negation words

We consider two ways to express opposite meanings. The first is *antonyms*, e.g., "good" vs. "bad"; and the second is through *negation* words, e.g., "good" and "not good".

We relied on the lists of antonyms provided by the MOE dictionary (cf. Sect. 2.2.1). Since there are only 8625 words in the antonym lists in the dictionary, we can handle only a very small number of antonyms at this moment.

We created a list of negation words based on our own judgment. This list of negation words include "無", "未", "不", "非", and "沒有". Note that we consider "無", "未", "不", and "非" to be negation words only when they are individual words after segmentation. Hence, we will handle words like "並非" correctly. This list allows us to find that statements (5) and (6) (used in NTCIR-10 RITE-2) have opposite meanings.

(5) 千禧年危機俗稱Ｙ２Ｋ危機或千禧蟲
(6) 千禧年危機並非Ｙ２Ｋ危機或千禧蟲

We could also handle other negation words like "無法", "未能", "不行", and "不能". However, this heuristic list is as yet unable to handle all possible Chinese negation words correctly. A more complex word like, "無可厚非", would need special attention in our system. A direct application of our heuristic list will treat this word as two negations, but this word is not really related to negation.

---

[14] http://en.wikipedia.org/wiki/S%C3%B8rensen%E2%80%93Dice_coefficient

### 2.2.3 Named entity and verb recognition

Among parts of speech in almost all languages, nouns and verbs are the essential parts for understanding the core meanings of sentences. Information about named entities such as persons, locations, organizations, and time are crucial for inferring relationships between statements. A software tool for named entity recognition (NER) not only annotates words in a sentence as nouns but also subcategorizes them as persons, locations, organization names and time specifications. Although current technologies for NER do not offer perfect performance, being able to carry out NER even partially paves a way to handle typical questions regarding the five Ws (What, When, Where, Why, Who). We employed S-MSRSeg, which is a tool for named entities recognition developed by Microsoft Research (Gao, Li, Wu, & Huang 2005).

Verbs provide information about the actions or states that a given sentence describes. Recognizing verbs for a sentence pair is thus useful. We employed the Stanford parser (Levy & Manning 2003) to do the tagging of parts of speech. Although it is possible to consider sub-categorization of verbs, we did not do so in the current study.

### 2.3 Syntactic features

We parsed the Chinese statements with the Stanford parser (Levy & Manning 2003) to obtain the parse trees and the part-of-speech (POS) tags for words. A parse tree of a sentence reveals important information about the meaning of the sentence. At this moment, we used the parsing results to do two types of comparisons. The first was to compare the similarity between the parse trees of *T* and *H* with the same method (the DICE coefficient) that we used to compare the definition trees of different senses as explained in Sect. 2.1.1. We also compared the collections of POS tags of two sentences, particularly the tags for verbs.

Based on our experience, the Stanford parser works better for simplified Chinese than for traditional Chinese. Hence, we converted statements of traditional Chinese into simplified Chinese before the parsing step in our procedures (cf. Sect. 2.1.1).

We noticed that the Stanford parser did not always produce the best or even correct parse trees for the given statements. The parser ranked candidate parse trees with probabilistic models, and produced the trees with leading scores. Although we could request more than one parse tree for a given statement, we chose to use only the top-ranked tree for computational efficiency of our systems.

### 2.4 Semantic features

It is preferable to employ higher level information about statement pairs to judge their entailment relationships. After considering information available at the lexical and syntactic levels, semantic features immediately came to mind. However, there are multiple ways to define and represent sentential semantics. Frame semantics is a conceivable choice (Fillmore

page 8/36

Table 1 Matrix form for encoding dependency structures

|         | We | consider | dependency | structures | inferring | textual | entailment |
|---------|----|----------|------------|------------|-----------|---------|------------|
| We      | 0  | **1**    | 0          | 0          | 0         | 0       | 0          |
| consider| 0  | 0        | 0          | 0          | 0         | 0       | 0          |
| dependency | 0 | 0      | 0          | **1**      | 0         | 0       | 0          |
| structures | 0 | **1**  | 0          | 0          | 0         | 0       | 0          |
| inferring | 0 | **1**   | 0          | 0          | 0         | 0       | 0          |
| textual | 0  | 0        | 0          | 0          | 0         | 0       | **1**      |
| entailment | 0 | 0      | 0          | 0          | **1**     | 0       | 0          |

1976; Burchardt, Pennacchiotti, Thater, & Pinkal 2009), for instance. In this work, we explored an application of dependency structures (Chang, Tseng, Jurafsky, & Manning 2009).

Linguists consider the context of words a very important factor to define meaning. "You shall know a word by the company it keeps" (Firth 1957) or similar arguments (e.g., Firth 1935; Harris 1954) are commonly cited in courses on linguistics. "One sense per discourse, one sense per collocation" (Yarowsky 1995) appears in the literature in computational linguistics very frequently. For this reason, using vector space models to capture contextual information has become one of the standard approaches in both natural language processing and information retrieval.

In our work, we explored an application of dependency structures to capturing the contextual information in a sentence. There are different ways to apply the dependency structures for inferring entailment relationships, and we note that Day et al. also employed the tree-edit distances of dependency structures in NTCIR-10 RITE-2 (Day et al. 2013).

We illustrate our methods with a short English example, "**We consider dependency structures for inferring textual entailment**", to make the example more easily understandable to non-Chinese speakers. We list the typed and collapsed dependencies of this statement below. A dependency relation is expressed in the format of **relation-name(governor, dependent)**, where both governor and dependent are words appended with their positions in the sentence.

> nsubj(consider-2, We-1)
> root(ROOT-0, consider-2)
> amod(structures-4, dependency-3)
> dobj(consider-2, structures-4)
> prepc_for(consider-2, inferring-6)
> amod(entailment-8, textual-7)
> dobj(inferring-6, entailment-8)

We can ignore the root node and build a matrix to encode the direct relationships between words, as shown in Table 1. The column headings show the governors, and the row headings show the dependents. A cell will be 1 if there is a relationship from the dependent to the governor. Hence, ignoring the relation name, the cell (We, consider) is 1 because of nsub(consider-2, We-1). Notice that the matrix is not symmetric because of the functions of words in different relationships.

The matrix, denoted by *R*, encodes the holistic relationships between words in a statement, and can be considered a way to represent the context of words in a given statement. There are many similar applications of such matrices in computer science, e.g., for modeling connectivity between web pages (Page, Brin, Motwani, & Winograd 1998) and for modeling traffic networks (Liu & Pai 2006).

As *R* encodes only the direct relationships between words, we can compute the powers of *R* to explore the indirect relationships between the words. For example, a "1" in the second power of *R*, $R^2$, shows that there is a one-step indirect relationship between two words. If we compute the second power of the matrix in Table 1, we will find that the cell with "dependency" as the row heading and with "consider" in the column heading is 1—suggesting the idea of "consider dependency" in the statement. When we compute higher powers of *R*, we will find fewer "1"s in the matrices because there are fewer word pairs with very remote indirect relationships.

Based on such observations, we explored the possibility of encoding the sentential context with the union of the powers of *R* for a statement. In the reported experiments in this paper, we chose to compute the *XR* matrix, defined in Eq. (2), for a given statement. A cell in *XR* will be 1 if the cell at the corresponding positions in any of the first five powers of *R* is 1.

$$XR = R \cup R^2 \cup R^3 \cup R^4 \cup R^5 \qquad (2)$$

## 3 Classification Methods

Although machine learning-based algorithms are the most conceivable method for classification problems including the recognition of textual entailment (Dagan, Dolan, Magnini, & Roth 2009), the size of training data available at NTCIR-10 RITE-2 was not large enough to make us feel comfortable to just take this intuitive avenue. Hence, in addition to applying support vector machines, we also tried to come up with our own parameterized heuristic functions to make classification decisions. The parameters would be tuned with the training data, so, technically, we can still consider our first approach as a machine-learning based method.

### 3.1 Trained heuristic functions

We explain the individual factors that we considered in our heuristic function in the following section.

**Table 24.** Effects of considering syntactic and semantic information indecisive

|  | Traditional Chinese | | | |  | Simplified Chinese | | | |
|---|---|---|---|---|---|---|---|---|---|
|  | RITE.Test | | RITE-2.Test | | | RITE.Test | | RITE-2.Test | |
|  | MacroF1 | Acc. | MacroF1 | Acc. |  | MacroF1 | Acc. | MacroF1 | Acc. |
| LM-5 | 71.50 | 71.89 | 64.60 | 64.81 | LM-11 | 71.58 | 77.64 | 62.16 | 65.94 |
| LM-5A | 71.70 | 72.00 | 64.51 | 64.81 | LM-11A | 70.95 | 77.15 | 62.31 | 65.81 |
| LM-6 | 72.48 | 72.89 | 64.10 | 64.36 | LM-12 | 72.36 | 78.13 | 62.05 | 65.81 |
| LM-6A | 71.32 | 71.67 | 64.34 | 64.70 | LM-12A | 69.25 | 75.68 | 62.03 | 65.56 |

TC were better than the best performing team which actually participated in NTCIR-9 RITE. Moreover, the accuracy achieved by LM-12 was also slightly better than the best accuracy for SC in NTCIR-9 RITE.

### 4.7 Effects of syntactic and semantic information

In order to study the effects of considering parse trees (F8 in Table 2) and the dependency structures (F16 in Table 2), we intentionally removed F8 and F16 from LM-5 and LM-6 in Table 14 and LM-11 and LM-12 in Table 15. We used LM-5A, LM-6A, LM-11A, and LM-12A to denote these new settings. Table 24 lists the MacroF1 and accuracy scores when we used LM-5A, LM-6A, LM-11A, and LM-12A with linearly weighted models to predict entailment.

Although we hoped that considering higher level linguistic information could make a significant contribution to the scores, the data does not support our hypothesis decisively. Most of the time, considering F8 and F16 made the classification results only relatively and marginally better for simplified Chinese. The effects of considering F8 and F16 were quite arbitrary for test data of traditional Chinese, as indicated by the left side of Table 24.

## 5 Additional Discussions

In this section, we discuss some issues that involve observations obtained in multiple experiments. More specifically, we discuss the implication that was suggested by the experiments reported in Sect. 4. Although one might expect that some approaches should have achieved better performance than others, such expectations might not be realized in the current study. We investigate the issues and elaborate on possible reasons for the gap between the actual results and expected outcomes in this section.

### 5.1 Y-precision, Y-recall, N-precision, and N-recall

Although we have focused mostly on the effects of using different methods and features on the achieved MacroF1 and accuracy scores, the values of the Y-precision, Y-recall, N-precision, and N-recall are informative for the design of algorithms.

Table 25. Performance statistics of teams which participated in both SC and TC subtasks in NTCIR-10 RITE-2

| Teams | Simplified Chinese | | | Traditional Chinese | | |
|---|---|---|---|---|---|---|
| | Ranks | MacroF1 | Acc. | Ranks | MacroF1 | Acc. |
| **JUNLP** | 24 | 48.49 | 48.66 | 16 | 48.72 | 48.81 |
| **IASL** | 10,18 | 60.45 | 63.25 | 1,14 | 67.14 | 67.76 |
| **MIG** | 2,6,12 | 68.09 | 68.50 | 2,3,6 | 67.07 | 67.54 |
| **CYUT** | 3,7,9 | 67.86 | 68.12 | 12,13,15 | 55.16 | 55.16 |
| **Yuntech** | 14,15,16 | 53.52 | 59.54 | 8,9,10 | 62.31 | 62.54 |
| **IMTKU** | 13,17,23 | 54.28 | 62.74 | 4,7,17 | 65.99 | 66.29 |
| **WHUTE** | 8,11 | 61.65 | 66.58 | 5 | 65.55 | 66.29 |

It should be noted that, when handling the statement pairs of simplified Chinese, our methods had high values in Y-recall and N-precision and low values in N-recall in Sects. 4.4 and 4.5. After using the training methods, our methods showed a tendency to grant entailed relationships to statement pairs. We suspect that this phenomenon may have resulted from the imbalanced portions of Y-pairs and N-pairs in the development set (cf. Table 3).

### 5.2 Performance of SVM-based systems

Indeed, it is not surprising that the quality of training data influenced the performance of the trained models. The amount of data available for training may have also affected the performances of teams which adopted supported vector machines (SVMs) as their classifiers. Table 25 shows some statistics of the performance of all of the teams which participated in the BC subtask for both simplified and traditional Chinese in NTCIR-10 RITE-2. Since each team could submit up to three runs of their systems, a team would have as many results as the runs they submitted. The "MacroF1" and "Acc." columns show the highest MacroF1 and accuracy achieved by the teams.

Among the seven teams, only IASL (Shih, Liu, Lee, & Hsu 2013) did not use SVMs, and MIG (Huang & Liu 2013) used SVMs in one of their three runs. The other five teams used SVMs as their classifiers, and only CYUT (Wu, Yang, Chen, Chiu, & Yang 2013) achieved better performance in simplified Chinese than in traditional Chinese. Although MIG's best performance in simplified Chinese is better than its best performance in traditional Chinese, as shown in Table 25, MIG's performance in simplified Chinese is actually poorer than its performance in traditional Chinese when MIG used an SVM-based classifier (cf. MIG-3 in Tables 5 and 6).

### 5.3 Effects of specific features on experiments with real test data

Comparing the experimental results discussed in Sects. 4.3, 4.4, and 4.5, we found that, overall, using systematic ways to search for parameters and features offered us more chances to achieve better performance than relying on results of intuitively selected experiments to build an inference system.

We have also attempted to compare many experimental results that were influenced by whether or not we considered synonyms in computing word overlap in Sect. 4. The following statement pair of NTCIR-10 RITE-2 provides an example of the need to consider synonyms. One needs to recognize the synonymous relationship between "聽力" and "聽覺" to correctly handle this pair.

(15) 噪聲對動物也有很大的影響，降低動物聽力
(16) 噪聲對動物的聽覺有很大的影響

Nevertheless, experimental results showed that considering synonyms only helped improve our performance in the TC experiments in NTCIR-10 RITE-2. Similar results were not observed in other experiments that we reported in Sects. 4.4 and 4.6. This may have resulted because the test data did not include many instances that really needed synonyms to make correct judgments and may have also been caused by imperfect judgment of synonymous relationships between Chinese words, which remains a very challenging problem for Chinese.

The entailment relationships between a statement pair may hold for a wide variety of reasons and their combinations, and the organizers of evaluation tasks try to cover as many different types of entailment relationships as possible in the datasets (Dagan, Dolan, Magnini, & Roth 2009; Shima et al. 2012; Watanabe et al. 2013). As a consequence, the overall performance might not be improved instantly due to the consideration of just one specific factor. Researchers have studied the correlation between datasets and performance of systems (Lin, Lee, Shih, & Hsu 2015). Hence, it may not be easy to single out and justify the extract contribution of a specific feature with real test data.

The same phenomenon occurred again when we tried to examine the effects of considering syntactic and semantic information to judge entailment relationships with experiments reported in Sect. 4.7.

## 5.4 World knowledge and subjective judgments

In the real world, we may not be able to judge whether one statement entails another solely by linguistic information (Vanderwende, Menezes, & Snow 2006; Dagan, Dolan, Magnini, & Roth 2009). This is particularly true when world knowledge, connotation and subjective judgments are involved. Following are some statement pairs that were used in NTCIR-10 RITE-2.

Knowledge about the conversion between "米" (meter) and "釐米" (centimeter) is required to judge whether (17) entails (18).

(17) 阿諾爾特大花草直徑能夠到達 3 米
(18) 阿諾爾特大花草直徑 50～90 釐米

The standard answer to the statement pair (19) and (20) is yes, probably because the annotator believed that something that is "最高" (highest) must also be "高" (high). However, this may not be always true, just like the best performer in a contest might not really achieve very high scores.

(19) 鹿茸滋補藥效最高

(20) 鹿茸藥效高

# 6   Concluding Remarks

The main goal of this paper is not to provide a comprehensive survey of studies on textual entailment. Rather, we provide empirical experience obtained from experiments with real test data in NTCIR-9 RITE and NTCIR-10 RITE-2. For additional survey articles that we have not discussed, readers might want to refer to (Androutsopoulos & Malakasiotis 2010) and (Watanabe el al. 2012).

In this paper, we presented the linguistic features and the computational models which we used to achieve second positions in the BC subtask for both simplified and traditional Chinese in NTCIR-10 RITE-2. Significantly extended investigations were carried out, reported, and analyzed to share our empirical experience in textual entailment based on the real data used in NTCIR-9 RITE and NTCIR-10 RITE-2. More experiments, including experiments on English test data used in PASCAL RTE-1 and RTE-2, are available in (Huang 2013).

Based on the experience and discussions reported in this paper, we believe that more work on true natural language understanding is needed to achieve better performance in textual entailment recognition. For future work, we are exploring the possibility of applying techniques of textual entailment for answering questions in reading comprehension tests that are designed for language learners (Huang, Lin, & Liu 2013). When computers can do the reading comprehension tests reasonably well, they might also explain the answers to students and serve as a learning companion.

## Acknowledgments

This research was supported in part by the student travel fund of the Department of Computer Science of National Chengchi University and in part by funding from the Grants of NSC-100-2221-E-004-014, NSC-101-2221-E-004-018, NSC-102-2420-H-001-006-MY2, and MOST-103-2918-I-004-001 of the Ministry of Science and Technology of Taiwan. Access to the digital library services of the Harvard Library was granted to the second author during his visit to Harvard University.